\documentclass[a4paper, 10pt, conference]{ieeeconf}      
\usepackage{FG2018}

\FGfinalcopy 

\IEEEoverridecommandlockouts                              
\usepackage{graphicx}
\usepackage{hyperref}
\usepackage{amsmath,amsfonts,amssymb}
\usepackage{booktabs} 
\usepackage{multirow}

\usepackage{color}

\def\FGPaperID{0153} 

\title{\LARGE \bf
Accurate Facial Parts Localization and Deep Learning for 3D Facial Expression Recognition
}

\author{\parbox{16cm}{\centering
    {\large Asim Jan$^1$, Huaxiong Ding$^2$, Hongying Meng$^1$, Liming Chen$^2$, Huibin Li$^3$}\\
    {\normalsize
    $^1$Department of Electronic and Computer Engineering, Brunel University London, UK\\
    $^2$LIRIS laboratory UMR CNRS 5205, Ecole Centrale de Lyon, France\\
    $^3$School of Mathematics and Statistics, Xi'an Jiaotong University, China}
    }}

\begin{document}

\ifFGfinal
\thispagestyle{empty}
\pagestyle{empty}
\else
\author{Anonymous FG 2018 submission\\ Paper ID \FGPaperID \\}
\pagestyle{plain}
\fi
\maketitle

\begin{abstract}
Meaningful facial parts can convey key cues for both facial action unit detection and expression prediction.
Textured 3D face scan can provide both detailed 3D geometric shape and 2D texture appearance cues of the face
which are beneficial for Facial Expression Recognition (FER).
However, accurate facial parts extraction as well as their fusion are challenging tasks.
In this paper, a novel system for 3D FER is designed based on accurate facial parts extraction and deep feature fusion of facial parts.
In particular, each textured 3D face scan is firstly represented as a 2D texture map and a depth map with one-to-one dense correspondence.
Then, the facial parts of both texture map and depth map are extracted using a novel 4-stage process consists of
facial landmark localization, facial rotation correction, facial resizing, facial parts bounding box extraction and post-processing procedures.
Finally, deep fusion Convolutional Neural Networks (CNNs) features of all facial parts
are learned from both texture maps and depth maps, respectively and nonlinear SVMs are used for expression prediction.
Experiments are conducted on the BU-3DFE database, demonstrating the effectiveness of combing different facial parts, texture and depth cues and
reporting the state-of-the-art results in comparison with all existing methods under the same setting.
\end{abstract}

\section{INTRODUCTION}

Facial expressions are important tools used to communicate the emotional reaction and/or state of a person during their daily activities.
There are many expressions a human can display, and behind each emotion there are a group of components. These are the person's intentions, action tendencies, appraisals, other cognitions, neuromuscular and physiological changes, expressive behavior, and subjective feelings~\cite{Cohn2007}. These components cause the movement of the facial muscles which in return creates a visual expression for others to see the emotion.

3D imaging instruments provide the ability to capture all the muscle movement in an accurate way, regardless of the lighting and pose variations. From these high resolution 3D facial scans, the muscle activities are visually obvious, which is beneficial for facial expression recognition. However, there are two main challenges produced by it: 1) Normal 3D data is a point cloud in the 3D space. Automatic location detection of the frontal face and the registration with its associated 2D face image is very complex and difficult. 2) For facial expression recognition, different muscle activations have different impact on the shape of the face. The Facial Action Coding System (FACS)~\cite{Ekman1976} has defined the relationship between action units and the emotional state. However, the accuracy of current action unit detection is not high. The accuracy of action unit based emotion detection system is not high either.

In this paper, we focus on the task of  recognizing the six basic facial expressions by using both 2D appearance and 3D geometric shape cues.
Moreover, the importance of different facial parts will be fully explored.
In particular, a novel system for 3D FER is designed based on accurate facial parts extraction techniques
and deep CNN feature fusion schemes related to different facial parts.
To accurately corp meaningful facial parts from both facial texture maps and depth maps, we propose a novel 4-stage procedures consists of
facial landmark localization, facial rotation correction, facial resizing, facial parts bounding box extraction and post-processing steps.
To deeply explore the importance of different facial parts for FER, we propose a novel deep feature fusion sub-net which can efficiently
learn the importance weights of different CNN features associated with different facial parts.
The proposed system is evaluated on the public dataset and achieving the best results among other methods in the same setting.
The main contributions of this paper are the following:
\begin{itemize}
\item A novel accurate facial parts localization method is developed based on the 2D face alignment
techniques and the one-to-one dense correspondence prior information between facial texture maps and depth maps.
\item A novel deep fusion CNN subnet is designed to learn the combination relations and importance weights of different facial parts represented
by the pre-trained deep CNN features.
%
\item State-of-the-art performance (i.e. 88.54\% average FER rate) is achieved in comparison with all existing methods on the same dataset (i.e. BU-3DFE) and under the same protocol.
\end{itemize}

The rest of the paper is organized as follow. Related works are reviewed in section 2, and the details of the proposed approach is described in section 3.
In Section 4, we report the experimental results and section 5 concludes the paper.

\section{RELATED WORKS}


3D FER has become an extensive field of research with many early attempts in \cite{Wang2006,Soyel2007,gong2009automatic,berretti2010set,Lemaire2013} and most recent works in~\cite{Li2015,li2017multimodal,Jan15} that trend to use both 2D and 3D multi-modal data to further improve the accuracy.
Huynh et al.~\cite{Huynh2016} proposed to use deep CNNs for classifying the six basic facial expressions.
Two CNNs are trained on the BU-3DFE database based on the 2D facial appearance and the the 3D face shape, respectively.
Li et al.~\cite{Li2015} thoroughly investigates FER using both 2D and 3D modality, starting with hand-crafted descriptors to capture local shape and local texture information. This work was followed by introducing deep learning techniques \cite{li2017multimodal} in which they achieved state-of-the-art performances on the BU-3DFE database.
Recently, deep learning is becoming an emerging solution, of which different forms are utilized for certain tasks to aid 2D FER. A popular idea that has been utilized for emotion detection and recognition is the fast representation of the Deep Belief Networks (DBNs). This was developed by Hinton et al. \cite{Hinton2006} using a greedy algorithm to quickly learn a generative model one layer at a time. Applications for DBNs include trying learn and understand the facial expression behaviors in \cite{Lv2014,Liu2014}. However, which regions of the face can provide better discrimination towards facial expressions has not been fully explored in these systems.


Lv et al.~\cite{Lv2014} proposed a framework that has multiple learning stages to try and understand the face and parts of the face. They use a DBN as an hierarchical detector that starts of by looking for the face, which is achieved by using a sliding window technique to capture HOG features. The detected face then has patches taken from it, and with this, they use the same approach to detect parts of the face. This continues until smaller individual parts are detected. Zhong et al.~\cite{Zhong2015} splits the face into small non-overlapping patches from which they try to categorize groups of patches. These categories are based on the relation between the different expressions, which include the common facial patches, specific facial patches, and the rest. They found that only having the highly discriminative patches improves performance, and that having too many patches causes a decrease with volatility in performance. Essentially, including too many patches introduces noise that makes the task more challenging. Most recent works by Li et al.~\cite{li2017eac} propose to use deep CNNs to learn regions of interest across the faces. These regions are detected using landmarks paired with an enhancing subnet, and the active regions are cropped using a CNN trained for cropping. These regions are then trained to detect for facial action unit activations in the form of muscle movements.

The mentioned works propose ideas that look deeper into the facial structure, to analyze patches instead for better discrimination of expressions. The mentioned systems have interesting ideas that can be exploited and improved using recent and advanced face detection, alignment and localization technologies~\cite{Yu2016,Asthana2014,Ren2014,Zhang2010}. They can be utilized to provide a consistent and less noisy solution. Facial parts such as the Mouth, Eyes, Nose and Eyebrows can be accurately captured and extracted. They can provide the samples to a framework that can analyze what facial parts work better and for which expression.

Having small patches as proposed by Zhong et al.~\cite{Zhong2015} lacks the ability to automatically localize and determine what part is in effect for an expression. It can also have trouble distinguishing patches when applied on faces of different gender and ethnicity. Having a learned detector for the face and parts as proposed by Lv et al.~\cite{Lv2014} cannot guarantee the robustness as much as face detection, alignment and localization techniques do, especially when there are peculiar samples provided. Both proposals used the JAFFE and CK+ database, which are not very diverse when it comes to their subjects ethnicities. In their experiments, they adopt a cross-validation approach, in which they do not ensure a subject independent protocol. This can result in a high performance which can be inconsistent when a new subject is ever tested.

This paper looks to create a robust face alignment procedure to accurately crop out meaningful 2D and 3D facial parts for facial expression recognition.
Meanwhile, these cropped facial parts are described by a set of deep CNNs features and their importance weights are fully explored by learning a deep feature fusion CNN subnet. Finally, the discriminative deep fused features of multiple facial parts is paired with a multi-class SVM for facial expression prediction.




\section{Proposed Approach}
\subsection{Framework Overview}

\begin{figure*}[t]
     \centering
       \includegraphics[width=16cm]{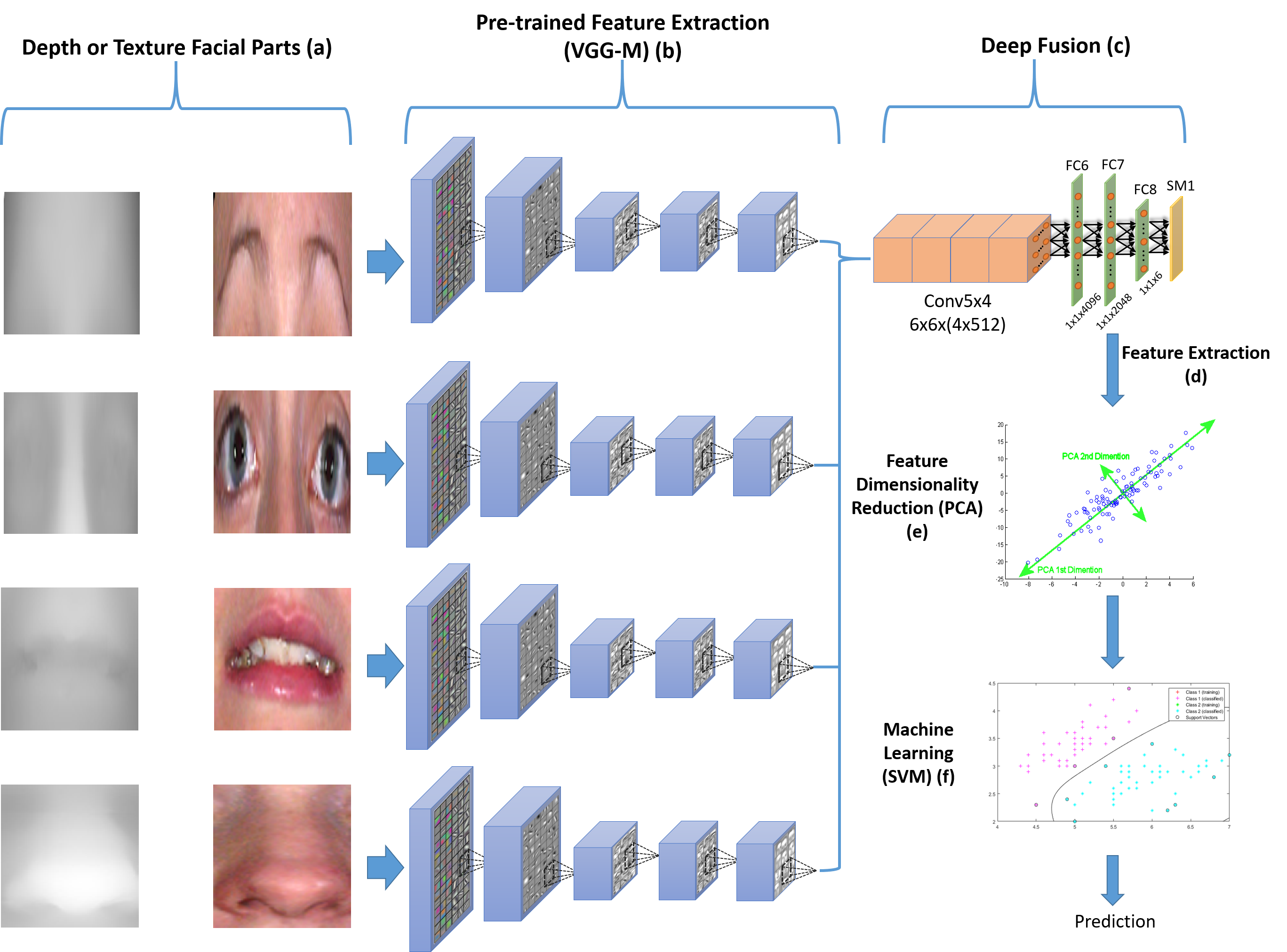}
      \caption{Framework designed to learn facial expressions. Starting with (a), facial parts are generated for both depth and texture images and trained separately. (b) Deep features are extracted from each facial part using the pre-trained VGG-NET-M \cite{Chatfield14} CNN. The features are taken after the Conv-5 layer, which is pooled to a size of $6\times6\times512$ per facial part. The feature maps for each facial part are fused together producing a higher dimensional feature map of $6\times6\times2048$. (c) is the fine-tuning process that is applied directly on the feature maps that learns the relation between the facial parts, by interconnecting them through a series of fully connected layers. Stochastic Gradient Descent is used for back-propagation, with the SoftMax layer as the loss function. Once the fine-tuning layers are trained for the depth and texture images separately, the FC7 layer features ($2\times2048$ dimensions) are extracted for each sample in (d). They are reduced in dimensionality (e) using PCA, keeping 99\% variance. Finally in (f), a SVM model with a polynomial kernel is adopted to learn and predict the facial expressions.}
      \label{fig:jointArch}
\end{figure*}

The proposed framework for FER using both 2D and 3D facial parts is illustrated in Fig.~\ref{fig:jointArch}.
Given textured 3D face scans, we first generate well-aligned texture and depth images as in~\cite{Li2015}.
Then, 2D facial landmarks are detected and used to correct facial pose and extract 4 key facial parts (i.e. Eyebrows, Eyes, Mouth and Nose).
Once facial parts are extracted from the texture and depth images, they are resized and propagated individually through a pre-trained CNN and
the resulting feature at layer FC7 is extracted as the deep representation of the facial part. In this case, a pre-trained network is preferred to a newly trained network due to the limited 3D facial data samples. The extracted deep features for each facial part are fused together through a smaller deep fusion network, that focuses on learning and interconnecting the contributions of each facial parts for FER. Once the parameters of the deep fusion network are learned, fused and fine-tuned deep features are extracted and SVM is used for the final expression classification.
Deep features are also captured from the whole face, but without the need of the fine-tuning part (c). This is to compare the performance of using
whole face against the facial parts. The details of the whole framework are introduced as follows. 

\subsection{Facial Parts Extraction}

To extract the facial parts from a sample, an approach using facial localizing and correction techniques is undergone. This is to obtain accurate and consistent facial parts. This process is broken down into 4 stages. 
The initial stage involves obtaining the 2D texture and depth maps from the textured 3D facial scans, which is followed by using facial landmark localization on to the 2D textured faces to generate 49 facial landmarks. Using these facial landmarks, the second stage corrects the rotation of each face to be fixed at a $0$ degree angle. The third stage resizes all the faces within the image spatial dimensions to provide consistency across all the samples. And finally, the fourth stage creates bounding boxes for each facial part using the normalized facial landmarks and the normalized texture and depth images of the face. These are cropped out and resized to have spatial dimensions of $64\times64$ pixels. Since facial texture image and depth image have been well-aligned, all the following 4 stages conducted on the texture image can also be applied to the
depth image.

\subsubsection{Facial Landmark localization}

\begin{figure}[t]
     \centering
      \includegraphics[height = 8cm]{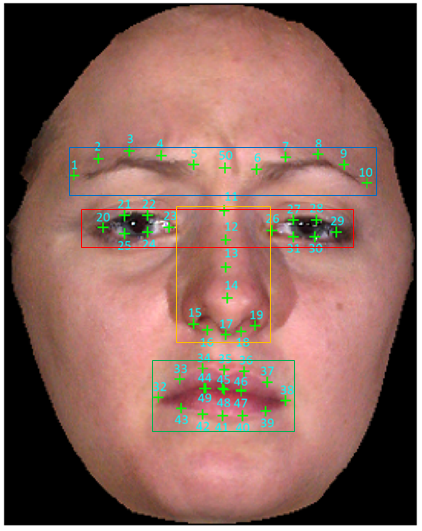}
      \caption{A sample face with the facial pre-processing applied. Facial landmarks are annotated on the face, along with the number for each point and bounding boxes containing each facial part.}
      \label{fig:LocalizedFace}
\end{figure}

2D texture face images are created by projecting each textured 3D facial scan into a 2D regular grid domain using the mapping and interpolation techniques in~\cite{Li2015}. Following this, 49 2D facial landmarks are generated on the 2D texture face images using a protocol called Incremental Parallel Cascade of Linear Regressors (iPar-CLR \cite{Asthana2014}). These facial landmarks are able to be transferred to the 3D textured face space as the one-to-one correspondence between 3D and 2D textures can be preserved during the projection mapping. The facial landmarks for the 3D face space also have a one-to-one correspondence to the 3D geometry of the face, making the alignment between the two the same. The 2D depth maps as well as the transferred landmarks are captured by projecting the landmark aligned textured 3D face scans into the same 2D regular grid domain as 2D texture face images. The iPar-CLR technique that is used for the landmark generation is based on the Supervised Descent Method \cite{asthana2013robust,xiong2013supervised}, which has been updated for parallelization and allow incremental updates when given new samples. The outcome from this technique produces a set of facial landmarks $P=\{P_{1},P_{2},...,P_{49}\}$ for each facial image as shown in Fig. \ref{fig:LocalizedFace}, where each point can be represented as an x and y coordinate $P_{i}=x_{i},y_{i}$. These facial landmarks will be utilized by the remaining 3 stages to correct the face and locate the facial parts.

\subsubsection{Facial Rotation Correction}

The rotation correction is achieved by calculating and reducing the angle between two lines. The first line $l_1$ goes vertically up the face within the image, starting from a central point close to the middle of the face, to a point that reaches the top of the forehead. The second line $l_2$ is based on the same central point, but goes vertically up the spatial image at 0 degrees rather than the face. The angle between the two lines determine how much the face is rotated. To produce the first line, two facial landmarks can be used to produce a straight line that cuts the face. These are: the point at top of the mouth ($P_{35}$) as the central point, and a point that can be generated in-between the inner eyebrows points $P_{5}$ and $P_{6}$. As most of the face is symmetrical across a vertical plane, the chosen facial landmarks provide a good reference point across all the samples. $P_{50}$ represents the point in-between the eyebrows ($P_{5}$ and $P_{6}$), which is calculated in eq.~\ref{eq:p50}.

\begin{equation}
\begin{aligned}[l]
P_{50}=(x_{50},y_{50})=\bigg(\frac{x_{6}-x_{5} }{2},\frac{x_{6}-x_{5} }{2}\bigg)
\label{eq:p50}
\end{aligned}
\end{equation}

The angle $\alpha$ can then be calculated between line $l_1$ and the image norm $l_2$ following eq. \ref{eq:V},~\ref{eq:Z} and \ref{eq:S}. Once the angle $\alpha$ is calculated, the whole image is rotated around point $P_{35}$ to make $\alpha=0\deg$.

\begin{equation}
\begin{aligned}[l]
l_1=(x_{50}-x_{35},y_{50}-y_{35})
\label{eq:V}
\end{aligned}
\end{equation}
\begin{equation}
\begin{aligned}[l]
l_2=(0,y_{50}-y_{35})
\label{eq:Z}
\end{aligned}
\end{equation}
\begin{equation}
\begin{aligned}[l]
\alpha=\cos^{-1}\bigg(\frac{l_1^{T}l_2}{||l_1||\times||l_2||}\bigg)
\label{eq:S}
\end{aligned}
\end{equation}

\subsubsection{Facial Resizing}

Once all the samples are rotated to the 0 degrees, the next step is to resize the faces to be approximately the same size, whilst retaining the original images spatial dimensions. This is a crucial step to make sure that the facial parts across expressions are of the exact same scale. Resizing the face to match the same size across all of the samples can be tricky. It cannot be guaranteed between different subjects as their face shape may vary. However, it should be possible to achieve for each expression from the same subject. This is done by measuring a distance between two points on the face that do not move between expressions. Then, if all the following samples are resized to produce the same distance between the same points, the face should ideally become the same size.
Two points that can be used are the inner eye corners $P_{23}$ and $P_{26}$. This is preferred over the inner eyebrows points $P_{5}$ and $P_{6}$ as the eyes are more towards the center of the face. The euclidean distance $d_{23,26}$ in eq.~\ref{eq:d2635} measures the distance between both inner eye points.
	
\begin{equation}
\begin{aligned}[l]
d_{23,26}= \sqrt{\big(x_{26}-x_{23} \big)^2+\big(y_{26}-y_{23} \big)^2}
\label{eq:d2635}
\end{aligned}
\end{equation}

Initially, a reference distance for the inner eye distance is required that can be used on all the samples. This can be determined by taking the mean value of a random batch of faces. Each sample can then have its face scaled bigger or smaller until its distance matches the reference distance.

\subsubsection{Facial Parts Extraction}

Once the face are aligned and resized correctly, the facial parts can be extracted. In this paper, four key facial parts including the
Eyebrows, Eyes, Mouth and Nose are considered. This is achieved by creating a bounding box around each facial part using the relevant facial landmarks, as demonstrated in Fig. \ref{fig:LocalizedFace}. The size of the walls for each bounding box is increased by approximately 5-10 pixels, to ensure that each facial part is fully inside each bounding box. Each image is cropped with its respective bounding boxes to separate the facial parts, and producing four new images per sample.

\begin{figure}[t]
     \centering
      \includegraphics[width = 5.5cm]{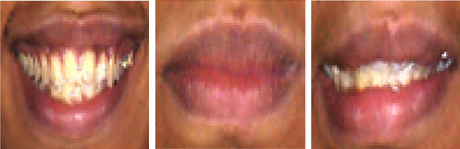}
      \caption{Extracted mouth parts with happy, neutral and fear expressions.}
      \label{fig:facialParts}
\end{figure}

\subsubsection{Facial Parts Post-Processing Steps}

Not all facial parts are the same in size across all the samples. Therefore, another processing step is taken to resize all the parts into $64\times64\times3$ images. This is to provide consistency in size for the hand-crafted and deep learning algorithms. The image resize should not alter the facial parts differently to other samples, as the initial face resizing that has taken place is done on the face within the image spatial dimensions. Fig.~\ref{fig:facialParts} shows sample images of the Mouth after all the processing techniques. The Happy, Neutral and Fear expressions are presented from the same subject.

\subsection{Fusion of Deep Facial Parts Features}



The next stage extracts deep features from the facial parts and a deep fusion network is deployed to learn the relations between each facial part. CNNs is used to train and extract deep features from the facial part images. An existing pre-trained CNN (VGG-M) is exploited for extracting robust features of the whole face and facial parts. These features are further trained through a fine-tuning process, learning a deep fusion network to interconnect these pre-trained features with feedback using the facial expression information.

The deep fusion network is made up of a sequence of convolution, ReLU and pooling layers. The design approach is different to the DF-CNN \cite{li2017multimodal}, by individually learning the depth and texture images. This way is chosen to get a deeper understanding of the two modalities, which can then be fused at a later stage using the SVM classifier. The initial convolution layer FC6 connects all of the pre-trained features $6\times6\times512\times4$ (4 facial parts) into a $1\times1\times4096$ dense layer. This layer has the job of fusing the various facial parts together into a single 4096 dimensional representation. This is then followed by a ReLU layer, and another dense layer of $1\times1\times2048$ (FC7). This will provide a compressed feature representation of the fused facial parts that can be utilized for classification. Finally, there is a $1\times1\times6$ layer to provide the prediction for the deep fusion network, from which a SoftMax layer is used to generate the loss for back-propagation. The deep fusion network feature is denoted as VGG-M-DF, and the pre-trained VGG-M feature is denoted as VGG-M-FC7.

\subsection{Expression Classification}
In this paper, SVM with a polynomial kernel is used for classification as it is a popular and reliable method adopted in many computer vision applications. To make the task simpler and faster, Principal Component Analysis is applied to reduce the dimensionality whilst retaining 99\% of the feature variance.

\section{Experimental Results}

The following experiments are based on understanding how facial parts can be utilized for better discrimination between facial expressions.
Deep learning techniques will also be implemented to learn from the face and the extracted facial parts. This is to study and apply recent technologies and advancements of machine learning, demonstrating how they can process facial expressions.

In addition, we have also extracted two hand-crafted features for comparison.
In particular, the Histograms of Oriented Gradients (HOG)~\cite{2006DalalHOG} and
Uniform Local Binary Patterns (ULBP)~\cite{2002OjalaLBP} features are extracted
from each facial part, as well as the whole face, from both texture and depth images.
For the deep learning based tests that use the whole face or just a single facial part, the framework that simply uses a single branch in parts (a) and (b) in Fig. \ref{fig:jointArch}. The deep fusion part (c) will have only have a single feature map set ($6\times6\times(1\times512)$ at the Conv5 layer. For the hand-crafted feature tests, the hand-crafted features are extracted directly from the images in part (a), and parts (b),(c) and (d) are skipped. Early feature fusion is applied to all the hand-crafted features from each facial part, along with the PCA for dimensionality reduction. Finally, the nonlinear SVM classifier is used for the expression classification.

\subsection{BU-3DFE Database}

The BU-3DFE database \cite{yin2008high} was created by a research group from Binghamton University. The BU-3DFE database
contains a total of 2500 samples made up from 100 subjects. For each subject, there are 4 samples for each of the 6 basic expressions
 (i.e. Angry, Disgust, Fear, Happy, Sad and Surprise), along with a single sample of the subject's neutral face. The 4 samples per expression represent the levels of intensity for that expression, which goes from mild to strong.

\subsection{Experimental Protocol and Parameters Setting}

The experiments are implemented using a publicly available deep learning toolbox called
MatConvNet~\cite{Vedaldi2014}. For fair comparison, the widely used protocol is used.
 It involves running 100 tests using 60 random subjects out of the 100 available, with the highest 2 intensities for each of the 6 prototypical expression. That comes to a total of 720 samples per facial modality for each test. Each test will use 10-fold cross-validation in a subject independent manner (no sample from a subject in the training set will be found in the testing set). SVM with a polynomial kernel is set up for multi-class classification. The remaining 40 subjects are used for the fine-tuning process as in \cite{li2017multimodal}. For the network training of the fine-tuning process, the hyper-parameters are set as follows:
\begin{itemize}
\item Batch Size = 12;
\item Learning rate = 0.0002 which slowly decreases to 0.00002 over 150 epochs;
\item Momentum = 0.9;
\item Weight Decay (regularization) = 0.0001;
\item Nesterov’s Momentum Update \cite{Sutskever2013} is used;
\item Stochastic Gradient Descent is used as the learning optimizer;
\end{itemize}
With the following pre-processing techniques applied:
\begin{itemize}
\item Mean image is subtracted across all samples only when the Conv5 feature is extracted;
\item Data is augmented to flip each image horizontally;
\item Network weights are randomly initialized using Xavier's improved method \cite{He2016}.
\end{itemize}
After the deep fusion subnet training completes 150 epochs, the epoch in which to retain the network parameters will be based on the lowest validation error. If there are multiple lowest validation errors, then the higher epoch is selected.


\subsection{Experimental Results}

The experiment in Table~\ref{tab:WholeFace} is based on using the framework on the facial parts and the whole face. Hand-crafted features are also tested on the facial parts and the whole face, for comparison with the deep learning approaches. The tests are based on features that are extracted from the texture images, depth images, with a following set of tests on the fusion of both sets of features. 

Based on the results, starting with the tests on the whole face, the best performance is produced using the deep fusion network VGG-M-DF when fusing both texture and depth cues. The worst performance is produced using the VGG-M-FC7 feature without any subnet for fine-tuning, demonstrating how the deep CNN features can be improved when the parameters are further tuned for the application. When evaluating the depth cues of the whole face, the hand-crafted features have shown better performances compared to the texture cues. The depth cues can be considered to do a better job in highlighting the facial deformations that occur within the face compared to the texture cues. However, for the deep features, this had the opposite effect where the texture cues has yielded better results. This may be due to the depth images only containing a single channel, whereas the pre-trained networks are trained on 3 channels that are the RGB channels. The quality of the extracted deep feature will not be its best due to the lack of color information.

The results for the facial parts show a big improvement over using the whole face, for both deep learning and hand-crafted approaches. Nearly all the different feature sets have had an increase in performance switching to facial parts, with the best performing feature as VGG-M-DF on the facial parts, using the texture and depth fusion to achieve a performance of 88.54\% recognition rate. The biggest improvement is of 5.9\% when comparing ULBP on the whole texture image of the face and facial parts. 

In all the tests with all features, there is a clear indication that fusing the texture and depth cues provides significant improvements. This demonstrates that they have characteristics that complement each other.

\begin{table}[t!]
\centering
\caption{Average Recognition Rate of 100 tests using 10-fold cross-validation for the hand-crafted and deep features extracted from the whole face and the fused facial parts.}
\vspace{0.2cm}
\begin{tabular}{|l||c||c||c||c|}
\hline
\multicolumn{2}{|c|}{\textbf{Feature}}	&\textbf{Texture} &\textbf{Depth}  &\textbf{Both} \\
\hline
\multirow{4}{0.8cm}{Whole Face}	&VGG-M-FC7	&77.96\%	&77.43\% &81.21\%\\
\cline{2-5}
	&VGG-M-DF	&81.77\%	&79.91\% &85.58\%\\
\cline{2-5}
	&HOG		&79.79\%	&81.10\% &84.04\%\\
\cline{2-5}
	&ULBP		&78.49\%	&78.80\% &81.89\%\\
    \cline{2-5}
\hline
\multirow{4}{0.8cm}{Facial Parts}	&VGG-M-FC7	& 83.76\%	& 75.72\% & 84.40\%\\
\cline{2-5}
	&VGG-M-DF	&\textbf{86.27\%}	&81.39\% &\textbf{88.54}\%\\
\cline{2-5}
	&HOG 		&85.03\%	&81.47\% &86.01\%\\
\cline{2-5}
	&ULBP 		&84.39\%	&\textbf{81.83\%} &86.89\%\\
\hline

\end{tabular}
\label{tab:WholeFace}
\end{table}

\begin{table}[ht]
\centering
\caption{Average Recognition Rate of 100 tests using 10-fold cross-validation for the deep features captured from the individual facial parts. }
\vspace{0.2cm}
\begin{tabular}{|l||c||c||c||c|}
\hline
{\textbf{Feature}}&\textbf{Facial Part}	&\textbf{Texture} &\textbf{Depth}  &\textbf{Both} \\
\hline
\multirow{4}{1.5cm}{VGG-M-DF}& Eyebrows	& 42.79\%	& 42.03\% &\ 45.50\%\\
\cline{2-5}
& Eyes		&50.25\%	&47.47\% &54.74\%\\
\cline{2-5}
& Mouth		&\textbf{78.71\%}	&\textbf{75.06\%} &\textbf{81.94\%}\\
\cline{2-5}
& Nose		&54.01\%	&49.03\% &56.90\%\\
\hline

\end{tabular}
\label{tab:SepFacialParts}
\end{table}

Table \ref{tab:SepFacialParts} is an evaluation of each individual facial part using the deep fusion framework. However, to test each facial part individually, Conv5 ignores the concatenation of the facial parts and just considers the VGG features from each individual facial part. The findings suggest the most expressive part of the face is the mouth, which is the facial part that can physically move the most. The texture cues have performed better than the depth cues, which is the consequence of the lack of color channels from the depth images. However, the fusion of both texture and depth representations has provided a significant boost for each facial part, with the best performance of 81.94\%.


\begin{table}[ht]
\centering
\caption{Comparison against other works based on averaging 100 tests of 10-fold cross-validation}
\vspace{0.2cm}
\begin{tabular}{|l||c||c||c|}
\hline
\textbf{Feature}	&\textbf{Domain}  &\textbf{Accuracy} \\
\hline

Berretti et al. \cite{berretti2010set}&3D	&77.54\%\\
\hline
Gong et al. \cite{gong2009automatic}	&3D	&76.22\% \\
\hline
Lemaire et al.	\cite{Lemaire2013}	&3D	&76.61\% \\
\hline
Soyel et al. \cite{soyel2007facial}	&3D	&67.52\% \\
\hline
Wang et al. \cite{Wang2006}& 3D	& 61.79\% \\
\hline
Yang et al. \cite{yang2015automatic} &3D	&84.80\%\\
\hline
Zeng et al.	\cite{zeng2013automatic}&3D	&70.93\%\\
\hline
Zhen et al. \cite{zhen2016muscular} &3D	&84.50\% \\
\hline
Li et al. \cite{Li2015}		&2D+3D	&86.32\%\\
\hline
Li et al. \cite{li2017multimodal}		&2D+3D	&86.86\%\\
\hline
ULBP Facial Parts		&2D+3D	&86.89\%\\
\hline
VGG-M-DF Facial Parts		&2D+3D	&\textbf{88.54\%}\\
\hline
\end{tabular}
\label{tab:Comparison}
\end{table}

Table \ref{tab:Comparison} shows the performance comparison against other state-of-the-art methods that follow the same strict and fair protocol. In terms of accuracy, the VGG-M-DF feature on the facial parts has provided a new state-of-the-art performance, with an increase of 1.86\% over the works by Li et al.~\cite{li2017multimodal}. When incorporating deep learning techniques, using a deep fusion subnet to learn the contributions of the facial parts has demonstrated that connections can be made between them at different stages, and not just from the raw images. When looking at the domains used by others, 3D has shown to be very popular. But when 3D is combined with 2D information, the best performances come from there.

\section{CONCLUSIONS AND FUTURE WORKS}

The works in this paper presents a novel strategy to automatically obtain accurate facial parts, and fuse them together in an effort to jointly learn their contributions through a deep fusion subnet. In general, the introduction of using isolated facial parts showed a clear-cut improvement over nearly every test taken when compared to using the whole face. The proposed deep fusion subnet showed a significant increase to the overall performance over hand-crafted techniques using the facial parts and the whole face.

For the hand-crafted features, the performance based on facial parts was also better than that produced by whole face. It demonstrates again that our proposed accurate facial parts extraction method is the key for performance improvement. Not only this, but the use of 2D and 3D information through texture and depth maps showed a consistent improvement when early fusion is applied via feature concatenation.

Based on these findings, further improvements can be made through the use of other face alignment and parts extraction techniques. 
Moreover, a better deep learning system can be designed to work with CNNs that are pre-trained for facial expression recognition.

\section{ACKNOWLEDGMENTS}

This work is partially supported by Royal Society and National Natural Science Foundation of China (NSFC) under cost share research exchange project
No. 61711530242. It was also partially supported by the French Research Agency, Agence Nationale de Recherche (ANR) through the Jemime Project under Grant ANR-13-CORD-0004-02 and the Partner University Fund (PUF) through the 4D Vision project.



\end{document}